# Deep Reinforcement One-Shot Learning for Artificially Intelligent Classification Systems

Anton Puzanov, Kobi Cohen

## Abstract

In recent years there has been a sharp rise in networking applications, in which significant events need to be classified but only a few training instances are available. These are known as cases of one-shot learning. Examples include analyzing network traffic under zero-day attacks, and computer vision tasks by sensor networks deployed in the field. To handle this challenging task, organizations often use human analysts to classify events under high uncertainty. Existing algorithms use a threshold-based mechanism to decide whether to classify an object automatically or send it to an analyst for deeper inspection. However, this approach leads to a significant waste of resources since it does not take the practical temporal constraints of system resources into account. Our contribution is threefold. First, we develop a novel Deep Reinforcement One-shot Learning (DeROL) framework to address this challenge. The basic idea of the DeROL algorithm is to train a deep-Q network to obtain a policy which is *oblivious to the unseen classes* in the testing data. Then, in real-time, DeROL maps the current state of the one-shot learning process to operational actions based on the trained deep-Q network, to maximize the objective function. Second, we develop the first open-source software for practical artificially intelligent one-shot classification systems with limited resources for the benefit of researchers in related fields. Third, we present an extensive experimental study using the OMNIGLOT dataset for computer vision tasks and the UNSW-NB15 dataset for intrusion detection tasks that demonstrates the versatility and efficiency of the DeROL framework.

The authors are with the Department of Electrical and Computer Engineering, Ben-Gurion University of the Negev, Israel. Email: poznov@post.bgu.ac.il, yakovsec@bgu.ac.il

This work was supported by the BGU Cyber Security Research Center.

# I. INTRODUCTION

In recent years there has been a sharp rise in networking applications, in which significant events need to be classified by processing large volumes of data under severe reliability constraints when only a few (or even no) training instances are available. Traditional classification methods often use hundreds to thousands or more training instances. Classifying events based on a small number (a few or even no) training instances is much more challenging, however, and is known as one-shot learning. To handle this challenging task, organizations often use human analysts to classify events with high uncertainty. However, it is crucial to limit the amount of data sent to human analysts given the system constraints (e.g., analysts' time, task load, etc.). Consider for example an intrusion detection system (IDS) implemented to detect cyber attacks. Once the system experiences a zero-day (i.e., previously unseen) attack, the IDS often sends the session summary records to a cybersecurity analyst for deeper event inspection, who in turn labels the records as malicious activity (or normal activity in the case of a false positive). A more challenging scenario occurs when normal activity changes rapidly and is hard to learn offline or due to a blur boundary between the observation of normal and abnormal behavior [1], [2]. Finally, computer vision tasks involving the deployment of camera sensor networks in the field require classifying events for civilian or military video surveillance [3]–[5]. When the fusion center cannot classify a relatively new event with a sufficiently high reliability based on the available information, it often sends the observed data to human inspectors for further inspection and labeling. The problem setting considered in this paper models these complex situations.

## A. Active Learning and One-Shot Learning for Classification Systems

Machine learning (ML) algorithms typically expect to have a great deal of data to learn from, which then becomes a serious limitation in environments that can only provide a limited number of training instances. Active Learning (AL) methods [6] focus on choosing the most influential samples to be labeled, and are often used in such scenarios. AL methods have been studied extensively in the past two decades and in particular have dealt with ways to predict the impact of each sample in the online AL setting. The problem considered in this paper is related to AL in the sense that the agent is required to decide whether to label a sample automatically or request the true label from an analyst. Since humans can learn new classes from a single example [7], we aimed to design an artificially intelligent (AI) agent that has similar abilities to avoid requesting classification labels too frequently. The challenge is more complex in the case of one-shot learning classification tasks, since a classifier must learn new classes from a single example (or few or none) [8]. In these cases, storage components could be designed judiciously [9].

While computer-based one-shot classification has attracted growing attention in recent years, the issue of how a system can improve performance under limited resources has not been successfully addressed. The algorithms currently used by organizations apply a threshold-based mechanism to decide whether to classify an object by a computer or send it to a human analyst for deeper inspection. However, this approach has two main drawbacks. It leads to a significant waste of resources since it does not take practical temporal constraints on system resources (e.g., analysts' time, task load, reliability, etc.) into account, and also requires judicious turning of the algorithm's parameters whenever the system setting is changed. In this paper, we propose a novel artificially intelligent framework to overcome these issues.

*B. Deep Reinforcement One-Shot Learning for Classification Systems*

*Our goal is to develop a holistic treatment of a real-world system for optimizing the operation of classifying significant events from a few or even no training instances based on both computing power and human analysts under limited resources. We aim at developing an AI one-shot classification algorithm that improves its performance online by learning from the human analysts' decisions (without the need for intervention for tuning) so as to maximize a system-wide objective function. We develop a novel Deep Reinforcement One-shot Learning (DeROL) framework to achieve this goal.*

Deep reinforcement learning (DRL) (or deep Q-learning) has attracted much attention in recent years due to its capability to provide a good approximation of the objective value (referred to as Q-value) while dealing with very large state and action spaces. In contrast to Q-learning methods that perform well for small-size models but perform poorly for large-scale models, DRL combines a deep neural network with Q-learning, referred to as Deep Q-Network (DQN), to overcome this issue. The DQN is used to map from states to actions in large-scale models to maximize the Q-value (for more details on DRL and related work see Sections I-D and II-C). In DeepMind's recently published Nature paper [10], a DRL algorithm was developed to teach computers how to play Atari games directly from the on-screen pixels, and reported strong performance in many games. In [11], the authors developed DRL algorithms for teaching multiple players how to communicate to maximize a shared utility. Strong performance was found for several players in MNIST games and the switch riddle. In recent years, there has been increased interest in using DRL methods in other fields. A survey of very recent studies can be found in [12].

Given the large state space and the partially observed nature of one-shot classification under temporal constraints on system resources considered here, developing a framework that incorporates DRL optimization methods into the design of the one-shot learning algorithm could provide effective solutions to real-world networking applications of classification tasks with limited resources. The proposed framework described here is dubbed Deep Reinforcement One-shot Learning (DeROL).

*C. Main Results*

Below, we summarize the main contributions:

1) **A novel deep reinforcement one-shot learning (DeROL) framework:** We developed a novel deep reinforcement one-shot learning (DeROL) framework for efficient management of the network resources. The novelty resides in incorporating DRL optimization methods in the design of a one-shot learning algorithm for artificially intelligent classification based on computing resources and human resources. Unlike other DRL frameworks that train the DQN by using many training instances from the same classes as in the testing phase, the basic idea behind the DeROL algorithm is to train the DQN to obtain a policy which is *oblivious to the unseen classes* in the testing data. In other words, DeROL aims to optimize an objective function with respect to the *belief vector* (i.e., a probability mass function over classes) regardless of the actual identity of the classes. Then, in real time, DeROL maps the current system state (e.g., the belief vector based on the one-shot learning process, the status of system resources, etc.) to operational actions based on the trained DQN to maximize the objective function. This paper is the first to suggest a holistic treatment of AI designs for one-shot classification using analysts, under real-world requirements, including one-shot learning, classification time, and analysts' resources. The framework handles a stream of samples, where each sample must be classified, either automatically or manually. We demonstrate that DeROL algorithm learns online to effectively trade off the exploration and exploitation of a classifier's prediction by either feeding it with manually labeled samples or accepting its predictions under the real world system constraints.

2) **Open source software:** We developed open-source software for the artificially intelligent classification system (see [13]). The open-source software is versatile and can be easily updated (see Section III-C for details). Our experimental study (discussed in the next paragraph) demonstrates the versatility of the software by successfully accomplishing two different challenging real-world classification tasks, one in computer vision, and the other in intrusion detection for cybersecurity.

3) **Experimental study:** We evaluated our framework using two different modern datasets. In the first experiment, one-shot computer vision tasks were examined using the popular OMNIGLOT dataset [7]. In this experiment, the DeROL algorithm needs to learn to classify an image into a matching class after only seeing a few training samples. In the second experiment, network traffic was analyzed to discover cyber attacks using the modern UNSW-NB15 intrusion detection dataset [14], [15]. In this experiment, network session summary records were classified to distinguish between attacks and normal network operations. To the best of our knowledge, this is the first paper to organize and test this dataset on a one-shot learning task. This experiment was conducted to examine the ability of the DeROL algorithm to adapt to rapid attack changes (e.g., zero-day attacks). In both experiments, the algorithm successfully demonstrated its ability to learn efficient resource management policies without the need for judicious tuning as is required in AL algorithms. Significant performance gains over the commonly used Least-Confidence Active-Learning algorithm [16] were observed.

### D. Related Work

The control component in our setting that decides whether to request labels or automatically classify samples is closely related to the field of active learning (AL). AL methods are typically used to train classifiers on a limited number of labeled examples by allowing the algorithm to choose which samples should be labeled or not [6]. AL algorithms usually take an iterative approach, where it is assumed that all the data are constantly available, and at each iteration the most informative samples are labeled. Selecting the most valuable samples is a key step in improving learning efficiency. This problem dates back to Chernoff's framework for optimal experimental design. In this setting, the decision maker is allowed to choose the experiment to be conducted at each time, where different experiments generate observations from different distributions under each hypothesis. The statistics literature has focused on maximizing the rate function in which the classification error decreases (see classic results in [17], and more recent studies in our previous work [18] and references therein). Traditional ML approaches focused on evaluating the least confidence score [6], where the samples with the lowest classification confidence are manually labeled, and margin sampling [16], where the difference between the two most confident samples is used to detect samples. More recent approaches [19] aim at weighting several features, such as uniqueness and similarity. In [9], the authors developed an algorithm that trains an RL agent to perform AL in an online manner. However, maximizing a system-wide objective function under practical temporal constraints on system resources as considered in this paper leads to fundamentally different settings and algorithm design.

The focus on classifying events from a small number of training instances is closely related to the one-shot learning problem. Despite recent advances and vast research in the field on ML, this problem has not been completely addressed. One-shot learning algorithms can be broadly divided into two groups. The first tries to generalize the knowledge obtained from past samples. In [7], the authors suggested a method which allows for reasoning about the obtained samples, and implemented it for images. Once an image is obtained, the algorithm breaks it down into meaningful components and enriches its knowledge by transforming and combining known components. In [20], the authors suggested a way to predict the Neural Network (NN) parameters from a single sample by training a teacher network. In [21], the authors suggested building a probabilistic model for every seen class. The second approach relies on memory components for storing the known samples, and teaching the algorithm to match the sample in question to the sample bank. In [8], this type of algorithm learns a general distance function using NN, and uses it to find a similarity between a given sample and the samples in the bank. In [22], the authors suggested a framework, in which the ML algorithm has access to a databank with read and write heads. This allows the algorithm to make use of historical samples, and uses Least Recently Used Access to reason about relevant information. In this paper, we employ the latter approach.

The AI framework developed here combines DRL optimization methods in the algorithm design to obtain good policies for online one-shot classification. DRL makes it possible to train AI agents by reward and punishment without specifying exactly how a task should be accomplished [23]. An RL agent was successfully trained to play various games, such as backgammon [24] and Atari [10] by training a DRL agent that takes actions by observing the raw pixels. This was considered as a great leap forward over traditional RL capabilities, and has been the basis for many other advances, such as using double Q-learning [25], in which two DQNs simultaneously learn the policy and value evaluation. An LSTM-based DRL agent was suggested in [26], and was used to solve problems where temporal relations need to be inferred, such as when an agent goes through a maze. Other DRL tasks have been discussed for combining DRL in Active-Learning [9], Web scrapping [27], Q-learning based robot control [28], and wireless networking management [29]–[35].

## II. System Model and Problem Formulation

### A. Description of the System Model

Consider a system that monitors a stream of events (e.g., image samples, network connection summary records, etc.) and needs to classify each event by arrival (up to a permitted delay). Samples might arrive from one-shot classes, which are relatively new classes with only a few (or even no) training instances. At each given time there are up to $K$ possible classes (which can change over time). To correctly classify the events, $M$ human analysts assist in the classification system and manually label the samples when asked. Let $a(t)$ be the action taken at time $t$, and let

$$\mathcal{A} \triangleq \{C_1, ..., C_K, A, D\} \tag{1}$$

be the action space. The description of the actions and their associated rewards are detailed next. A summary of the available actions and their associated rewards are given in Table I.

Action $a(t) = C_i$ is taken when the system *classifies* the event under category $i$ automatically without the help of the analyst at time $t$. A reward $R_C$ is received each time the classifier takes action $C_i$, $i = 1, ..., K$. When the classifier makes a correct classification, then $R_C = R_{\text{correct}}$, where $R_{\text{correct}} > 0$ is a positive constant. Otherwise, $R_C = R_{\text{wrong}}$, where $R_{\text{wrong}} < 0$ is a negative constant.

Action $a(t) = A$ is taken when the system sends the event sample to the *analyst* at time $t$ to request a label by performing a deeper inspection. Once a sample has been sent to analysts

TABLE I

Available Actions and Associated Rewards

| Action Description | Action Symbol | Reward |
|---|---|---|
| Automatic Classification | $C_i$ | $R_C$ |
| Sending to Analyst | $A$ | $R_A$ |
| Delaying Classification | $D$ | $R_D$ |

to request a label, it is assigned to a free analyst, and the true label is obtained after a random classification time required to inspect the sample and label it. We denote the analyst's random classification time by $T_A$ which follows an unknown distribution $f_{T_A}(\cdot)$. If there are no available analysts, the sample enters a FIFO queue and will be inspected in turn. Let $M_A(t) \in \{1, ..., M\}$ be the number of active analysts who are currently working on labeling samples. We denote the number of pending samples by $P(t)$. When there is at least one available analyst (i.e., $M_A(t) < M$), then $P(t) = 0$. Otherwise, if $M_A(t) = M$, then $P(t) \geq 0$. Given these definitions, we can define the analyst load by:

$$L_A(t) \triangleq \frac{M_A(t) + P(t)}{M}. \tag{2}$$

Note that $0 \leq L_A(t) < 1$ when there is at least one available analyst, and $L_A(t) \geq 1$ when all analysts are active. Furthermore, we take into account that the analysts have a buffer with a finite queue length to store samples, which is defined by $q_A$. A negative reward $R_A < 0$ is received each time the system takes action $A$, which is a monotonically deceasing negative function (when the queue is not full) with $L_A(t)$ (i.e., the price for sending samples to an analyst increases with the analyst load). If the queue is full, then the system discards the sample, and $R_A$ gets a constant negative reward (smaller than $R_{\mathbf{wrong}}$ to avoid these actions by the policy).

Action $a(t) = D$ is taken when the system *delays* the classification decision of the current sample. Intuitively, a good policy should delay samples when a similar event has been observed and sent to an analyst while its label is pending. Then, once a label for the similar sample is received, classifying the delayed sample with high reliability should be possible without sending it to an analyst at all, which allows the algorithm to judiciously manage system resources. Furthermore, we take into account that the buffer that stores delayed samples has a finite queue length, which is defined by $q_D$. Let $S_D(t) \in \{1, ..., q_D\}$ be the number of delayed samples at time $t$. Given these definitions, we can define the delayed sample load by:

$$L_D(t) \triangleq \frac{S_D(t)}{q_D},$$ 

(3)

where $0 \leq L_D(t) \leq 1$. Assume that the system delays a sample (say sample $x$), which in turns was stored in the delay buffer for $T_D(x)$ time units until it was classified. When $L_D(t) = 1$ (the buffer is full) the system discards the sample. A negative reward $R_D < 0$ is received when the system takes action $D$ which depends on $L_D(t)$ and $T_D$. If $L_D(t) = 1$ (i.e., the system discards the sample), then $R_D$ gets a constant negative reward (smaller than $R_{\mathbf{wrong}}$ to avoid these actions by the policy). Otherwise, $R_D < 0$ is a monotonically decreasing negative function with $T_D(\cdot)$ (i.e., the price for postponing classification decisions increases with the classification delay).

## B. The Objective

Let $r(t)$ denote the reward obtained by taking action $a(t)$ at time $t$. The reward depends on a complex partially observed environment with temporal correlations among actions as described in Section II-A. Let

$$R = \sum_{t=1}^{T} \gamma^{t-1} r(t)$$ 

(4)

be the accumulated discounted reward, where $0 \leq \gamma \leq 1$ is a discounted factor, and $T$ is the time horizon of the entire classification process of the total examined samples. We often set $\gamma = 1$, or $\gamma < 1$ when $T$ is bounded or unbounded, respectively. The objective is to find a policy $\phi$ that maximizes the expected accumulated discounted reward:

$$\max_{\phi} \quad \mathbf{E}\left[R(\phi)\right],$$ 

(5)

where $\mathbf{E}\left[R_n(\phi)\right]$ denotes the expected accumulated discounted reward when the system performs strategy $\phi$.

We are interested in developing a model-free learning algorithm to solve (5) that can effectively classify one-shot events under limited resources, while adapting to different network settings and requirements. Computing an optimal solution, however, is clearly intractable due to large state space and partial observability of the problem. Thus, we apply a DRL approach given its ability to provide good approximate solutions while dealing with very large state and action spaces.

In the next section we start by describing the basic ideas of Q-learning and DRL. We then develop the proposed algorithm that uses a deep reinforcement learning approach for one-shot classification design in Section III.

## C. Background on Q-learning and Deep Reinforcement Learning (DRL)

Q-learning is a reinforcement learning method that aims at finding good policies for dynamic programming problems. It has been widely applied in various decision making problems, primarily because of its ability to evaluate the expected utility from among available actions without requiring prior knowledge about the system model, and its ability to adapt when stochastic transitions occur [36]. The algorithm was originally designed for a single agent who interacts with a fully observable Markovian environment (in which convergence to the optimal solution is guaranteed under some regularity conditions in this case). It has been widely applied to more involved settings as well (e.g., multi-agent, non-Markovian environments) and demonstrated strong performance, although convergence to the optimal solution is open in general under these settings. Assume that we can encode the entire history of the process up to time $t$ to a state $s(t)$ which is observed by the system. By applying Q-learning to our setting, the algorithm updates a Q-value at each time $t$ for each action-state pair as follows:

$$
\begin{aligned}
Q_{t+1}\left(s(t), a(t)\right) = & \; Q_t\left(s(t), a(t)\right) \\
& + \alpha \left[ r(t+1) + \gamma \max_{a(t+1)} Q_t\left(s(t+1), a(t+1)\right) \right. \\
& \left. - Q_t\left(s(t), a(t)\right) \right],
\end{aligned}
\tag{6}
$$

where

$$
r(t+1) + \gamma \max_{a(t+1)} Q_t\left(s(t+1), a(t+1)\right)
\tag{7}
$$

is the learned value obtained by getting reward $r(t+1)$ after taking action $a(t)$ in state $s(t)$, moving to next state $s(t+1)$, and then taking action $a(t+1)$ that maximizes the future Q-value seen at the next state. The term $Q_t\left(s(t), a(t)\right)$ is the old learned value. Thus, the algorithm aims at minimizing the Time Difference (TD) error between the learned value and the current estimate value. The learning rate $\alpha$ is set to $0 \leq \alpha \leq 1$, which typically is set close to zero. Typically, Q-learning uses a sliding window to encode the recent history when the problem size is too large.

While Q-learning performs well when dealing with small action and state spaces, it becomes impractical when the problem size increases, mainly for two reasons: (i) A stored lookup table of $Q$-values for all possible state-action pairs is required, which makes the storage complexity intolerable for large-scale problems. (ii) As the state space increases, many states are rarely visited, which significantly decreases performance.

In recent years, DRL methods that combine deep neural network with Q-learning, referred to as Deep Q-Network (DQN), have shown great potential for overcoming these issues. Using DQN, the deep neural network maps from the (partially) observed state to an action, instead of storing a lookup table of Q-values. Furthermore, large-scale models can be represented by the deep neural network well so that the algorithm can preserve good performance for very large-scale models. Although convergence to the optimal solution of DRL is an open question, strong performance has been reported in various fields as compared to other approaches. A well known algorithm was presented in DeepMind's recently published Nature paper [10] for teaching computers how to play Atari games directly from the on-screen pixels. For other recent developments see Section I-D.

Due to the large state space and the partially observed nature of one-shot classification in various network application tasks, our Deep Reinforcement One-shot learning (DeROL) algorithm uses a DRL approach to the design of one-shot learning for artificially intelligent classification systems that operate under limited resources.

## III. THE DeROL FRAMEWORK FOR ARTIFICIALLY INTELLIGENT CLASSIFICATION SYSTEMS

A good policy should effectively collaborate with human analysts to maximize (5). It should take actions based judiciously on the observed state, while learning online to improve classification performance from a small amount of data that represents each class, and effectively adapt to different settings. In this section, we develop the DeROL framework to solve (5). The DeROL algorithm applies general large and complex settings and does not require prior knowledge on the underlying distribution of the process.

Section III-A presents the proposed architecture of the DeROL framework. Section III-B presents the pseudocode of the algorithm. Section III-C describes in detail the open-source software of the DeROL framework.

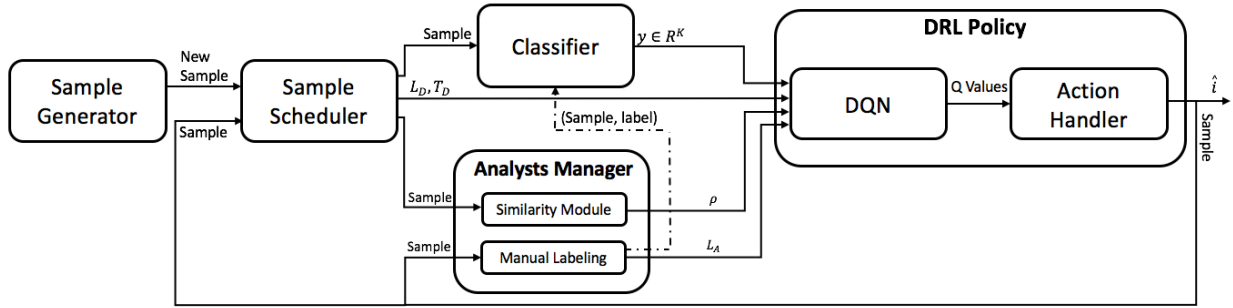

Fig. 1. An illustration of the DeROL architecture as described in Section III-A

## A. Architecture of the DeROL Algorithm

In this section, we describe the architecture of the DeROL framework. Our method is composed of two main components. The first is the classifier which is responsible for analyzing the samples from one-shot classes. The second is the DRL policy, which is responsible for taking good actions based on the classification output and the observed parameters from the system (e.g., analyst load, delay load, similarity of the sample to previous samples, etc.). *To address the challenge of obtaining good policies for versatile one-shot learning tasks under temporal correlations and limited resources, the key to the DeROL algorithm is to train the DRL policy independently of the identity of the training classes and identify the features required for efficient training.* Specifically, the policy uses input parameters such as the belief vector of the classes (produced by the classifier), and the similarity of the sample to previous samples. At the same time, it is oblivious to the actual identity of the classes. We will now describe in detail the main components of the algorithm. An illustration of the block diagram of the framework and the relations between the components is presented in Fig. 1.

*1) The Classifier:* Let $x$ be an event sample (e.g., an image matrix, a feature vector of network connection summary records, etc.), and $K$ be the maximal number of possible classes currently in the system (which can change over time). The classifier outputs soft decisions by mapping the input sample $x$ to $y(x) \in \mathbb{R}^K$, where each entry $y_i(x) \in y(x)$ holds a resemblance score between the input sample $x$ and class $i$ (i.e., a measure of the belief that $x$ belongs to class $i$). It is often the case that the mapping consists of several steps; e.g., preprocessing, normalization, and convolution with filters. In our framework, this logic is embedded in the classifier. The classifier must have one-shot learning capabilities. Note that in typical classification settings, one would require a classification output $i^*(x) = \arg\max_i y_i(x)$. However, since the DRL policy needs to maximize the expected accumulated reward, it uses soft decisions by the classifier as an input.

Setting the classifier separately from the policy has important advantages. First, it allows the framework to be versatile and easily deployed in different environments. This is achieved due to the decoupling of the overall logic from the specific domain representation. The classifier block can be updated easily as new versions are developed and optimized independently of the entire system. Second, a subset of the training samples can be taken solely for the purpose of training the classifier to reduce the noise in the system, produce more stable results, and improve the convergence speed. Third, the performance of the classifier can be examined independently of the policy to quickly diagnose issues in the system. Finally, this type of system can use pre-trained classifiers from legacy systems or use transfer learning, which has demonstrated excellent performance, by reusing neural networks.

One-shot classifiers usually have a memory component. In our framework, the one-shot classifier consists of a batch of stored label samples which have recently entered the system, and a distance function used to measure the similarity between the new sample and the stored ones. Let $\mathcal{W}$ be a set of recently labeled samples stored by the classifier, and let $\mathcal{W}_i \subseteq \mathcal{W}$, $i = 1, ..., K$, be a subset of the labeled samples from class $i$. To classify a new sample $x$, the distance between $x$ to each class (say $i$) is computed with respect to all the database samples in $\mathcal{W}_i$. A maximal distance of $d_{max}$ is used to truncate the computed distance and represents classes that are not present in the database. Specifically, the distance between sample $x$ to class $i$ is given by

$$d_i(x) = \min\left(d_{max}, \min_{z \in \mathcal{W}_i} d(z, x)\right), \tag{8}$$

where $d(z, x)$ is a distance function between samples $z$ and $x$ (i.e., nearest neighbor type distance), and can be chosen properly depending on the application. For example, in our experimental study (see Section IV), we used a Modified Hausdorff Distance for image classification, and the Euclidean Distance for network traffic analysis. Finally, to transform the distance into a similarity score, we took the inverse operation $y_i(x) = d_{max} - d_i(x)$, resulting in $K$ similarity scores for the $K$ classes. Transforming into a belief vector can be done mapping to $[0, 1]$ (e.g., using a soft-max function).

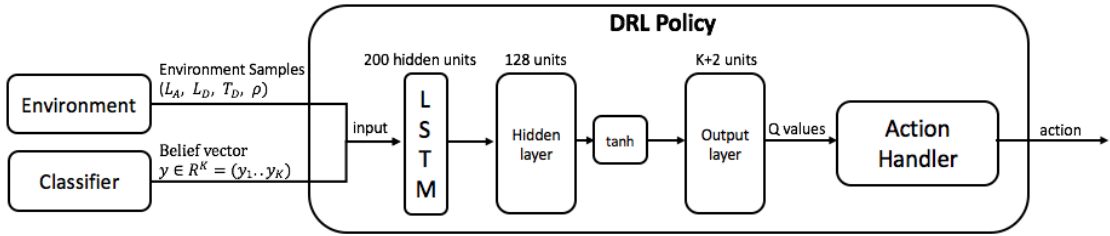

Fig. 2. An illustration of the DRL policy components as described in Section III-A.2.

*2) The DRL Policy:* We illustrate the DRL policy in Fig. 2. At each time $t$, when a sample (say $x$) has entered the system, the DRL receives the following parameters as input: (i) the classifier output $y(x) \in \mathbb{R}^K$; (ii) the analyst load $L_A(t)$; (iii) the delayed sample load $L_D(t)$; (iv) the total time the sample has been stored in the delay buffer, $T_D(x)$; and (v) the similarity score $\rho(x)$ that measures the similarity between $x$ and the previous samples that have been sent to the analysts but have not been labeled yet (handled by the Analyst Manager as discussed in Section III-A.3). We denote the input to the DRL policy (i.e., the observation) by $o(t)$:

$$o(t) \triangleq \{y(x), L_A(t), L_D(t), T_D(x), \rho(x)\}, \tag{9}$$

where sample $x$ is processed at time $t$.

The DRL policy first maps the observation $o(t)$ to the corresponding Q-values by a trained DQN. Note that training the DQN is oblivious to the actual classes and is only affected by the temporal relations between the beliefs about the classes and the system resources. This makes the DeROL framework highly suitable for one-shot learning. Since the policy must learn from temporal correlations, we employ the LSTM layer to extend the static classification context to the temporal one. The LSTM layer is then followed by a fully connected layer and the tanh activation function. We used the Adam optimizer [37] for training. The DQN outputs a vector of $K + 2$ Q-values, one value for each action $(C_1, ..., C_K, A, D)$. The Action Handler then takes the action with the highest Q-value (where we applied $\epsilon$-greedy policy for exploring actions).

*3) Other Components: Sample Generator, Sample Scheduler, and Analyst Manager:* The framework consists of several components, resulting in a modular architecture and logic encapsulation. The implementation details are discussed in Section III-C. Here, we describe the logic connections between the components. The Sample Generator reads samples from a given dataset and enters them into the system. The new samples together with the delayed samples flow to the Sample Scheduler, which consists of two queues, one for each sample type. The Sample Scheduler reports the delay buffer load $L_D(t)$. The Sample Scheduler outputs a single sample at each time-step to be processed by the framework, with the delayed samples having precedence over the new ones. The sample is examined by both the Classifier, which outputs the soft vector $y(x) \in \mathbb{R}^K$ described in Section III-A.1, and the Analyst Manager, which evaluates the similarity between the input sample and the entire samples currently being processed by the analysts. Samples marked for manual classification are processed by the Analyst Manager (i.e., sent to the analyst queue). Once the true label $i(x)$ is available, the sample $x$ and its label $i(x)$ are used to update the classifier.

### B. Pseudocode of the DeROL Algorithm

Next, we present the pseudocode of the algorithm.

---

**DeROL Algorithm**

---

1)  **for** iteration $i = 1, ..., R$ **do**

2)   **for** new sample = $m = 1, ..., n$ **do**

3)    add $m$ to sample scheduler ($\mathcal{S}$)

4)    **while** $\mathcal{S}$ has pending samples **do**

5)     get sample $s$ from $\mathcal{S}$

6)     form $o(t)$ following (9)

7)     Generate an estimation of the Q-values $Q(o(t))$ for available actions $a \in \{C_1, ..., A, D\}$ by the DQN

8)     Take action $a(t)$ according to policy $\phi$ given $Q(o(t))$

9)     obtain reward $r(t)$

10)    **end while**

11)   **end for**

12)   **if** Training Phase **do**

13)    Train DQN

14)   **end if**

15)   **end for**

---

*C.  Open-Source Software*

We developed an open-source software implementation of the DeROL framework. DeROL was developed using Python (see [13]). Fig. 3 illustrates the top level hierarchy of all the modules in the DeROL project, and in this section we summarize the implementation details. Interested readers are encouraged to read the README file for further information.

The main execution file is *train_derol.py*. The file contains the framework configurations, and is responsible for training and coordinating all the modules. Each package corresponds to a specific functionality of the framework and has different implementations as described below.

1) The *sample_generators* package contains the modules which are responsible for creating training and testing batches, as described by the Sample Generator block in Section III-A.3. Two sample generators have currently been implemented. The first is *OmniglotMulticlassGenerator*, which generates samples from the OMNIGLOT dataset [7]. It contains the logic for sampling, loading, pre-processing, and organizing the images. The second is *UNSWNormalizedTwoClass*, which is used to read, normalize, and organize network records from the UNSW dataset [14], [15]. More details on the OMNIGLOT and UNSW datasets are given in Section IV-A.

2) The *classifiers* package contains the classifiers implemented for the two classification tasks under DeROL. The classifiers are composed of a memory component and a distance metric, and correspond to the Classifier module described in Section III-A.3. Two classifiers are currently available, dubbed *OMNIGLOTClassifier* and *UNSWClassifier* for classification tasks of images and network records, respectively.

3) The *analyst_manager* package contains the related logic and information for the analysts, such as the number of total and free analysts, and is the counterpart of the Analyst Manager module described in Section III-A.3. It holds and manages the analysts queue, and produces the similarity metric $\rho(\cdot)$.

4) The *rl_agents* package contains the implementation of the RL agent in the system, as described in Section III-A.2.

Additional supporting modules include:

1) The *delay* package which contains generators for delay modeling. Currently, three types of generators have been implemented for modeling the delay: constant delay, uniformly distributed delay, and Poisson distributed delay.

2) The package *sample_handlers* is responsible for holding all the samples in the DeROL framework. It contains the *DelayedSamples* class for managing the samples that were delayed by the policy. This component is related to the Sample Scheduler component as described in Section III-A.3.

3) The package *trained_models* is not a module, but a storing location for all the trained agents. It is used for fault tolerance, backup, and experiments.

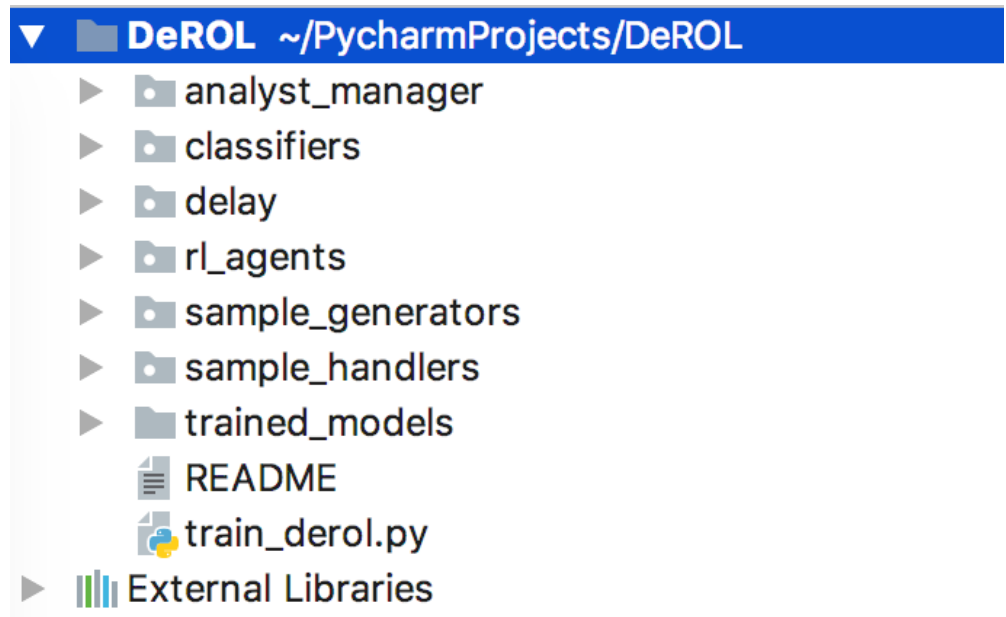

Fig. 3. The DeROL project folder as seen in PyCharm IDE. Each package has modules that implement different types of logic in the framework. The main execution file is train_derol.py.

## IV. EXPERIMENTAL STUDY

In this section we present an extensive experimental study to evaluate the DeROL algorithm. We examined DeROL performance under a one-shot learning setup, and used two different datasets to test its effectiveness in versatile environments. We compared the DeROL algorithm to the Least-Confidence Active-Learning (AL) algorithm [16] which is commonly used in real-world applications, and was adapted to online AL as suggested in [6], by manually tuning the classification's confidence threshold. Samples with low classification confidence (i.e., below the threshold) are sent to an analyst for manual classification. We added another threshold for delaying a sample. *The DeROL algorithm, however, avoids this exhaustive parameter tuning by self-adapting its policy to different configurations.*

### A. Datasets

We evaluated our framework on two modern datasets. The first is the OMNIGLOT dataset [7], which is commonly used to examine one-shot image classifiers, and the UNSW-NB15 intrusion detection dataset [14], [15] that was created by the IXIA PerfectStorm tool in the Cyber Range Lab of the Australian Centre for Cyber Security (ACCS), which simulates modern network traffic and cyber attacks. OMNIGLOT consists of letters taken from $50$ languages and a total of $1623$ classes. An example of several letters is shown in Fig. 4. Each letter has only $20$ different samples. Images were resized to $20 \times 20$; the pixel values are normalized values in [0.0,1.0]. We used 85% of the samples for training and the rest for testing. UNSW-NB15 consists of a hybrid of real modern normal activities and synthetic contemporary cyber attack behaviors. Each record is represented by $47$ features. It is labeled by a binary predicate (malicious or benign), and by the type of cyber attack for the malicious traffic, which is one of 9 simulated attacks. The full feature list and descriptions can be found in [14]. We used 6 attack classes for training and 6 for testing together with the normal traffic, i.e., 3 attack classes appeared in both experiments. Different samples were taken for the normal and the shared attack classes for training and testing.

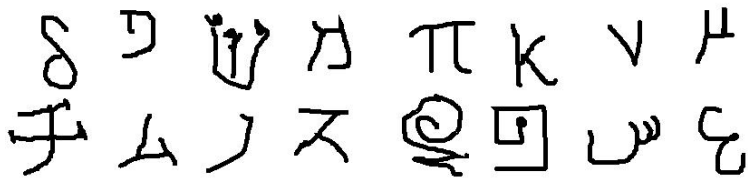

Fig. 4. An example of several letters from the OMNIGLOT dataset

### B. Computer Vision Classification Tasks using the OMNIGLOT Dataset

*1) Baseline Configuration:* We first evaluated DeROL performance on the OMNIGLOT dataset. We tested a total number $K = 30$ of classes. On each round, 3 classes were randomly selected. Each selected class had a random unknown label in the testing phase, which forced the classifier to request labels to learn online how to improve classification performance. Ten samples per class were randomly selected, resulting in a total of 30 samples which were then shuffled. These samples were sent to the sample scheduler and then sent to the agent. In the scheduler, the new samples were mixed with the delayed ones. The consumption of the new samples by the policy marked an end of a cycle. At this point, the statistics measures were collected and a new class was chosen and swapped with the older class in the generator. This environment forced one-shot classification since classes were changed rapidly. If the DRL policy delayed a sample, then the sample entered into a delay queue for 10 time-steps (and then was checked again). When a sample was sent to analysts, the inspection time was random and followed a distribution which was unknown to the policy. We modeled this by a uniform distribution with values $[1, 10]$. The number of analysts was set to $M = 3$. Each analyst worked on a single sample at any given time and the pending requests were held in a shared FIFO queue. Once an analyst finished inspecting a sample, it read the next sample from the queue or waited for new samples if the queue was empty. We set the maximal queue size for the delayed samples to $q_D = 100$ and the maximal queue size for manual classification to $q_A = 15$. If any of the queues was full, and the algorithm sent a sample to that queue, the sample was dropped and considered as a wrong classification. The reward for correct and wrong classification was set to $+1, -1$, respectively. The reward for sending a sample to an analyst was set to be a linearly decreasing function of the analyst load, $R_A = -0.1 \cdot L_A(t)$. The reward for delay of a sample was set to be an exponentially decreasing function of the delay, $R_D = -0.05 \cdot 2^{T_D/10}$.

*2) Results under the Baseline Configuration:* It is intuitive that good policies that learn how to manage system resources effectively are likely to generate more label requests for early instances of a class, achieve a higher correct classification rate as more instances of a class have been observed (i.e., improve the classification performance in an online manner), and balance the task loads for the analysts and the delayed samples effectively. Fig. 5.a, demonstrates the convergence of the DeROL algorithm to a good policy that significantly outperforms the Least Confidence AL algorithm in terms of the accumulated reward. Fig. 5.b, presents the empirical probabilities for each action as a function of the number of times that an instance from a class was observed.

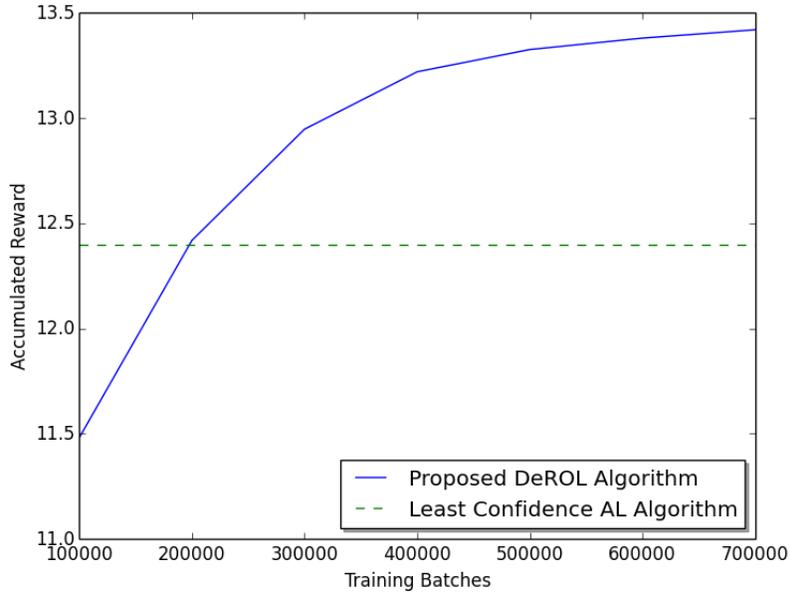

(a) Average accumulated reward

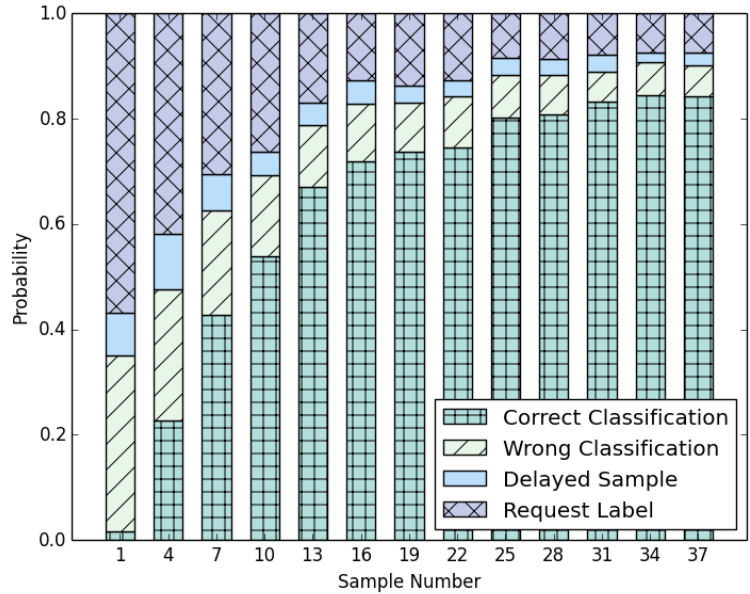

(b) Action probabilities as a function of the sample number

Fig. 5. Numerical results for the experiment described in Section IV-B.2. Fig. (a) presents the average accumulated reward under the DeROL algorithm and the Least Confidence AL algorithm. Fig. (b) shows the empirical probability of each action as a function of the number of samples observed for a class under the DeROL algorithm.

We denote by correct classification and wrong classification the cases in which action $C_i$ was taken and $i$ was the true class, or false class, respectively. We denote by *delayed sample*, and *request label* the cases in which actions $D$, and $A$ were taken, respectively.

Since the total number of samples in the system varies with the $x$-axis and backlogged delayed samples are presented in the figure, the actual correct classification rate of the samples (obtained by automatic classification plus analysts' inspections) is not derived from the figure. Therefore, we evaluated the actual correct classification rate below by counting the correct classifications (from the automatic classification plus the analysts' inspections) for each sample that entered the system:

$$\frac{\text{\# correct classification} + \text{\# request label}}{\text{\# total samples}} \approx 0.88. \tag{10}$$

It can be seen that in the first instances, the agent indeed requested more labels from the analysts, but that the later instances were correctly classified automatically by the classifier. These results demonstrate that the DeROL algorithm learns online to effectively trade off classification requests from the analysts and acceptance of the automatic classifier's outputs, with the goal of maximizing the accumulated reward.

*3) Increasing the Penalty for Wrong Classification:* We next tested our policy in a scenario in which wrong classifications incurs a very high penalty (specifically, we decreased the reward for wrong classification to $R_{\text{wrong}} = -5$). Thus, wrong declarations should be avoided as much as possible by increasing the rate of label requests and delayed samples (which will be correctly classified once similar samples return from the analysts). This scenario models typical situations found in medical or security applications

Fig. 6.a, presents the convergence of the DeROL algorithm to a good policy that significantly outperforms the Least Confidence AL algorithm in terms of the accumulated reward. Fig. 6.b, presents the empirical probabilities for each action as a function of the number of times that an instance from a class was observed. The actual correct classification rate below was obtained by counting the correct classifications (from the automatic classification plus analysts' inspections) for each sample that entered the system:

$$\frac{\text{\# correct classification} + \text{\# request label}}{\text{\# total samples}} \approx 0.94. \tag{11}$$

As shown, the classifier was much more conservative. It decreased the automatic classification rate significantly and fully utilized the analysts (the average analyst load was around $L_A \approx 0.97$),

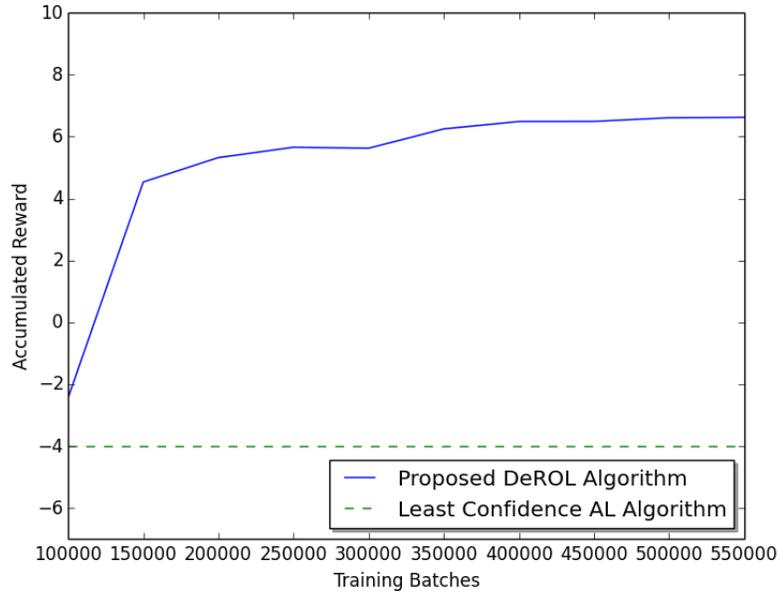

(a) Average accumulated reward

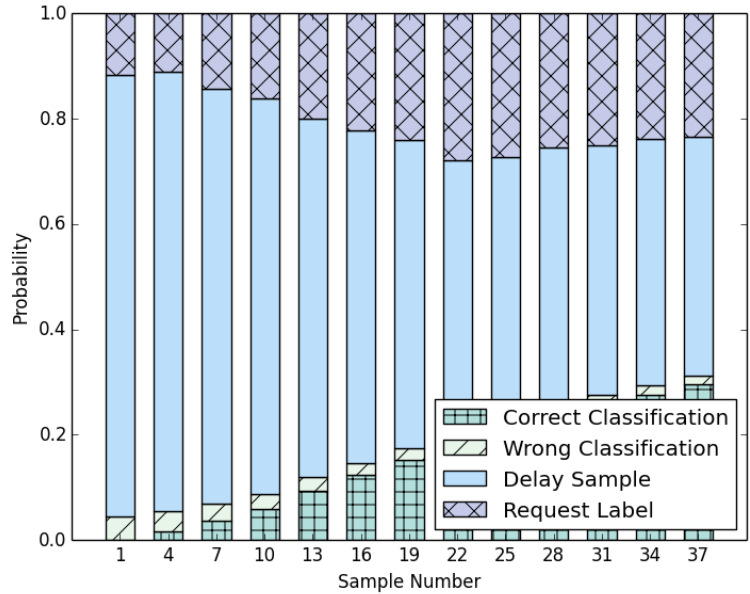

(b) Action probabilities as a function of the sample number

Fig. 6. Numerical results for the experiment described in Section IV-B.3. Fig. (a) presents the average accumulated reward under the DeROL algorithm and the Least Confidence AL algorithm. Fig. (b) shows the empirical probability of each action as a function of the number of samples observed for a class under the DeROL algorithm.

which in turn decreased the wrong classification rate (from $0.12$ in the previous experiment to $0.06$ in the current experiment). These results support our intuitive explanation for the adaptation of a good policy, and show that the DeROL algorithm successfully adapted to the change in the reward setting.

### C. Network Traffic Analysis using the UNSW-NB15 Intrusion Detection Dataset

The second dataset used for evaluation was the modern UNSW-NB15 intrusion detection dataset as described in Section IV-A. Our framework provides an effective solution to zero-day attacks previously unseen in the training data. Furthermore, unlike traditional intrusion detection methods that compare a network traffic record against an established baseline traffic (i.e., normal), our framework provides an effective solution to more challenging situations when the normal activity has a short history as well, and is hard to learn offline or due to a blur boundary between the observation of normal and abnormal behavior [1], [2]. Two changes in the modules of DeROL algorithm were made to analyze the UNSW-NB15 dataset as compared to the analysis of the OMNIGLOT dataset. First, the UNSW-NB15 dataset only contains 10 classes (normal plus 9 types of malicious activities). Since we needed to train DeROL to perform one-shot learning, we used a shorter classifier memory (i.e., 20 samples) equally divided between normal and malicious activity to effectively model the one-shot classification. Second, we used the Euclidean Distance for similarity checking of samples, since it performs better than the Modified Hausdorff Distance which is commonly used for image classification. This demonstrates the versatility of the DeROL algorithm to handle different types of classification tasks with simple adaptations.

We used the same parameters as in the first experiment except for the following changes. We tested a total number of 7 classes (normal plus 6 types of malicious activities). On each round, the sample generator entered 2 classes into the system, 20 samples from the normal class plus 10 samples from an abnormal class, keeping a majority of normal samples to model real-world scenarios, and capturing the one-shot effect. We changed the malicious class on each cycle. The consumption of the new 30 samples by the policy marked an end of a cycle. The number of analysts was set to $M = 2$. The reward for declaring an abnormal event as normal was set to $-30$, and the reward for declaring a normal event as abnormal was set to -2 (distinguishing between different abnormal classes was impractical due to the significant blur boundaries between abnormal classes, especially when using short memory for modeling the one-shot effect). The

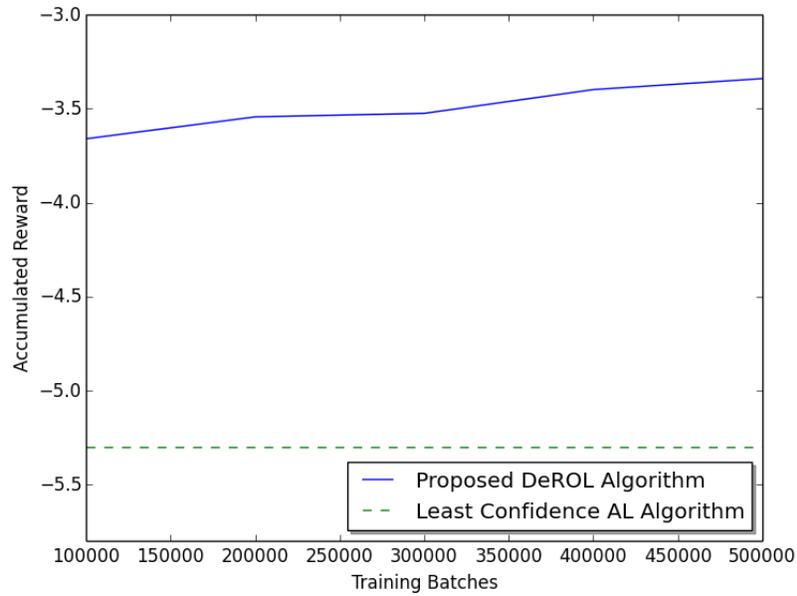

(a) Average accumulated reward

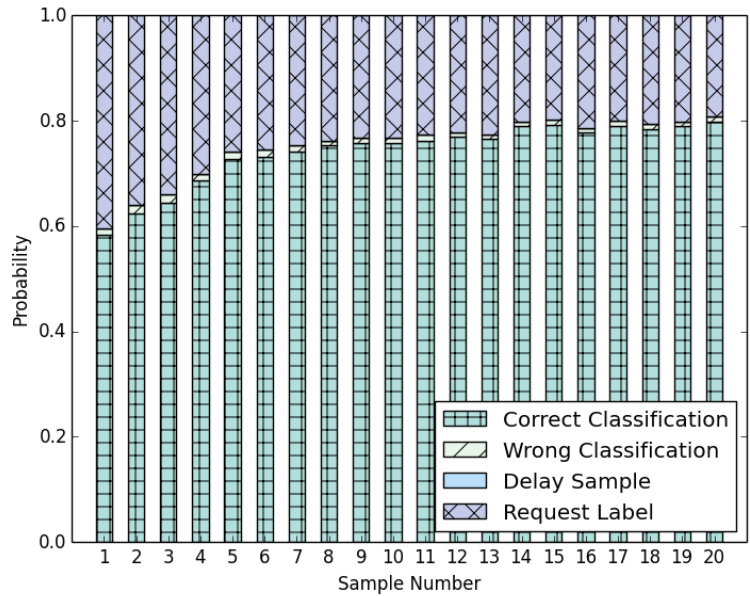

(b) Action probabilities as a function of the sample number

Fig. 7. Numerical results for the experiment described in Section IV-C. Fig. (a) presents the average accumulated reward under the DeROL algorithm and the Least Confidence AL algorithm. Fig. (b) shows the empirical probability of each action as a function of the number of samples observed for a class under the DeROL algorithm.

reward for correctly classifying normal, and abnormal events was set to 0, and 1, respectively. The reward for sending a sample to the analyst was set to be a linearly decreasing function of the analyst load, $R_A = -0.5 \cdot L_A(t)$. Since detecting cyber attacks typically cannot tolerate delay, we decreased the reward for delayed samples to $R_D = -2^{T_D/10}$ (i.e., we increased the penalty for declaration delays). Fig. 7.a presents the convergence of the DeROL algorithm to a good policy that significantly outperforms the Least Confidence AL algorithm in terms of the accumulated reward. Fig. 7.b presents the empirical probabilities for each action as a function of the number of times that an instance from a class was observed. The actual correct classification rate below was obtained by counting the correct classifications (from automatic classification plus analysts' inspections) for each sample that entered the system:

$$\frac{\# \text{ correct classification} + \# \text{ request label}}{\# \text{ total samples}} > 0.99. \tag{12}$$

These results show that the DeROL policy successfully adapted to the new environment, and achieved a very small error rate by effectively learning online from the analysts. As expected, the policy avoids delaying samples in this scenario, and makes decisions quickly. These results highlight the versatility of the DeROL framework, and its ability to effectively adapt to different configurations of one-shot classification tasks.

## V. Conclusion

We considered the problem of one-shot active learning classification in a real world environment. We developed a novel Deep Reinforcement One-shot Learning (DeROL) framework to address this challenge, and an open-source software for our implementation. We evaluated DeROL on two modern datasets, OMNIGLOT, which is commonly employed for testing one-shot image classification, and UNSW-NB15 which simulates a real-world application of malicious behavior classification in network traffic. We demonstrated that DeROL self-adapts to various system settings, achieves strong performance under different classification tasks using both datasets, and outperforms the commonly used Least Confidence AL algorithm.

## References

[1] M. N. Napiah, M. Y. I. B. Idris, R. Ramli, and I. Ahmedy, "Compression Header Analyzer Intrusion Detection System (CHA-IDS) for 6LoWPAN Communication Protocol," *IEEE Access*, vol. 6, pp. 16623–16638, 2018.

[2] E. Viegas, A. Santin, V. Abreu, and L. S. Oliveira, "Stream learning and anomaly-based intrusion detection in the adversarial settings," in *IEEE Symposium on Computers and Communications (ISCC)*, pp. 773–778, 2017.

[3] C.-Y. Chong and S. P. Kumar, "Sensor networks: evolution, opportunities, and challenges," *Proceedings of the IEEE*, vol. 91, no. 8, pp. 1247–1256, 2003.

[4] J. Xu, Y. Andrepoulos, Y. Xiao, and M. van Der Schaar, "Non-stationary resource allocation policies for delay-constrained video streaming: Application to video over internet-of-things-enabled networks," *IEEE Journal on Selected Areas in Communications*, vol. 32, no. 4, pp. 782–794, 2014.

[5] J. Ahn, J. Paek, and J. Ko, "Machine learning-based image classification for wireless camera sensor networks," in *IEEE 22nd International Conference on Embedded and Real-Time Computing Systems and Applications (RTCSA)*, pp. 103–103, 2016.

[6] B. Settles, "Active learning literature survey," Computer Sciences Technical Report 1648, University of Wisconsin–Madison, 2009.

[7] B. M. Lake, R. Salakhutdinov, and J. B. Tenenbaum, "Human-level concept learning through probabilistic program induction," *Science*, vol. 350, no. 6266, pp. 1332–1338, 2015.

[8] O. Vinyals, C. Blundell, T. Lillicrap, D. Wierstra, *et al.*, "Matching networks for one shot learning," in *Advances in Neural Information Processing Systems*, pp. 3630–3638, 2016.

[9] M. Woodward and C. Finn, "Active one-shot learning," *arXiv preprint arXiv:1702.06559*, 2017.

[10] V. Mnih, K. Kavukcuoglu, D. Silver, A. A. Rusu, J. Veness, M. G. Bellemare, A. Graves, M. Riedmiller, A. K. Fidjeland, G. Ostrovski, S. Petersen, C. Beattie, A. Sadik, I. Antonoglou, H. King, D. Kumaran, D. Wierstra, S. Legg, and D. Hassabis, "Human-level control through deep reinforcement learning," *Nature*, vol. 518, no. 7540, pp. 529–533, 2015.

[11] J. Foerster, Y. M. Assael, N. de Freitas, and S. Whiteson, "Learning to communicate with deep multi-agent reinforcement learning," in *Advances in Neural Information Processing Systems*, pp. 2137–2145, 2016.

[12] Y. Li, "Deep reinforcement learning: An overview," *arXiv preprint arXiv:1701.07274*, 2017.

[13] A. Puzanov and K. Cohen, "An open-source software to Deep Reinforcement One-Shot Learning (DeROL) Classification Framework," *available on GitHub: https://github.com/antonpuz/DeROL*.

[14] N. Moustafa and J. Slay, "Unsw-nb15: a comprehensive data set for network intrusion detection systems (unsw-nb15 network data set)," in *Military Communications and Information Systems Conference (MilCIS), 2015*, pp. 1–6, IEEE, 2015.

[15] N. Moustafa and J. Slay, "The evaluation of Network Anomaly Detection Systems: Statistical analysis of the UNSW-NB15 data set and the comparison with the KDD99 data set," *Information Security Journal: A Global Perspective*, pp. 1–14, 2016.

[16] K. Wang, D. Zhang, Y. Li, R. Zhang, and L. Lin, "Cost-effective active learning for deep image classification," *IEEE Transactions on Circuits and Systems for Video Technology*, vol. 27, no. 12, pp. 2591–2600, 2017.

[17] H. Chernoff, "Sequential design of experiments," *The Annals of Mathematical Statistics*, vol. 30, no. 3, pp. 755–770, 1959.

[18] K. Cohen and Q. Zhao, "Active hypothesis testing for anomaly detection," *IEEE Transactions on Information Theory*, vol. 61, no. 3, pp. 1432–1450, 2015.

[19] B. Demir and L. Bruzzone, "A novel active learning method in relevance feedback for content-based remote sensing image retrieval," *IEEE Transactions on Geoscience and Remote Sensing*, vol. 53, no. 5, pp. 2323–2334, 2015.

[20] L. Bertinetto, J. F. Henriques, J. Valmadre, P. Torr, and A. Vedaldi, "Learning feed-forward one-shot learners," in *Advances in Neural Information Processing Systems*, pp. 523–531, 2016.

[21] L. Fei-Fei, R. Fergus, and P. Perona, "One-shot learning of object categories," *IEEE transactions on pattern analysis and machine intelligence*, vol. 28, no. 4, pp. 594–611, 2006.

[22] A. Santoro, S. Bartunov, M. Botvinick, D. Wierstra, and T. Lillicrap, "Meta-learning with memory-augmented neural networks," in *International conference on machine learning*, pp. 1842–1850, 2016.

[23] L. P. Kaelbling, M. L. Littman, and A. W. Moore, "Reinforcement learning: A survey," *Journal of artificial intelligence research*, vol. 4, pp. 237–285, 1996.

[24] G. Tesauro, "Td-gammon: A self-teaching backgammon program," in *Applications of Neural Networks*, pp. 267–285, Springer, 1995.

[25] H. Van Hasselt, A. Guez, and D. Silver, "Deep reinforcement learning with double q-learning.," in *AAAI*, vol. 2, p. 5, Phoenix, AZ, 2016.

[26] B. Bakker, "Reinforcement learning with long short-term memory," in *Advances in neural information processing systems*, pp. 1475–1482, 2002.

[27] J. Rennie, A. McCallum, *et al.*, "Using reinforcement learning to spider the web efficiently," in *ICML*, vol. 99, pp. 335–343, 1999.

[28] S. Mahadevan and J. Connell, "Automatic programming of behavior-based robots using reinforcement learning," *Artificial intelligence*, vol. 55, no. 2-3, pp. 311–365, 1992.

[29] T. J. O'Shea and T. C. Clancy, "Deep reinforcement learning radio control and signal detection with kerlym, a gym rl agent," *arXiv preprint arXiv:1605.09221*, 2016.

[30] U. Challita, L. Dong, and W. Saad, "Proactive resource management in LTE-U systems: A deep learning perspective," *arXiv preprint arXiv:1702.07031*, 2017.

[31] S. Wang, H. Liu, P. H. Gomes, and B. Krishnamachari, "Deep reinforcement learning for dynamic multichannel access," in *International Conference on Computing, Networking and Communications (ICNC)*, 2017.

[32] S. Wang, H. Liu, P. H. Gomes, and B. Krishnamachari, "Deep reinforcement learning for dynamic multichannel access in wireless networks," *to appear in the IEEE Transactions on Cognitive Communications and Networking*, 2018.

[33] O. Naparstek and K. Cohen, "Deep multi-user reinforcement learning for dynamic spectrum access in multichannel wireless networks," in *IEEE Global Communications Conference (GLOBECOM)*, pp. 1–7, Dec 2017.

[34] O. Naparstek and K. Cohen, "Deep multi-user reinforcement learning for distributed dynamic spectrum access," *submitted to IEEE Transactions on Wireless Communications, arXiv preprint arXiv:1704.02613*, 2017.

[35] C. Zhang, P. Patras, and H. Haddadi, "Deep learning in mobile and wireless networking: A survey," *arXiv preprint arXiv:1803.04311*, 2018.

[36] C. J. Watkins and P. Dayan, "Q-learning," *Machine learning*, vol. 8, no. 3-4, pp. 279–292, 1992.

[37] D. P. Kingma and J. Ba, "Adam: A method for stochastic optimization," *arXiv preprint arXiv:1412.6980*, 2014.